\documentclass[10pt,twocolumn,letterpaper]{article}

\usepackage[pagenumbers]{cvpr} 
\usepackage[dvipsnames]{xcolor}

\definecolor{cvprblue}{rgb}{0.21,0.49,0.74}
\usepackage[pagebackref,breaklinks,colorlinks,citecolor=cvprblue]{hyperref}

\usepackage{multirow}
\usepackage{algorithm}
\usepackage{algorithmic}
\usepackage[utf8]{inputenc}
\usepackage{url}
\usepackage{booktabs}
\usepackage{amssymb}
\usepackage{bbding}
\usepackage{pifont}
\usepackage{wasysym}
\usepackage{utfsym} 
\usepackage{fontawesome}
\usepackage{multirow}

\usepackage{stfloats}
\usepackage{overpic}
\def\paperID{2900} 
\def\confName{CVPR}
\def\confYear{2024}

\newcommand\blfootnote[1]{%
  \begingroup
  \renewcommand\thefootnote{}\footnote{\begin{minipage}[t]{\linewidth}#1\end{minipage}}%
  \addtocounter{footnote}{-1}%
  \endgroup
}

\usepackage{pifont}
\newcommand{\corr}{\ding{61}}%

\title{GoMVS: Geometrically Consistent Cost Aggregation for Multi-View Stereo}

\author{Jiang Wu\textsuperscript{1 *}\; Rui Li\textsuperscript{1,2 *}\; Haofei Xu\textsuperscript{2,3}\; Wenxun Zhao\textsuperscript{1}\; Yu Zhu\textsuperscript{1 \corr}\; Jinqiu Sun\textsuperscript{1}\quad Yanning Zhang\textsuperscript{1\corr}\\
\textsuperscript{1}Northwestern Polytechnical University \ \textsuperscript{2}ETH Zürich \
\textsuperscript{3}University of Tübingen, Tübingen AI Center
}

\begin{document}
\twocolumn[{
\renewcommand\twocolumn[1][]{#1}
\maketitle
\begin{center}
    \captionsetup{type=figure}
    \includegraphics[scale=0.52]{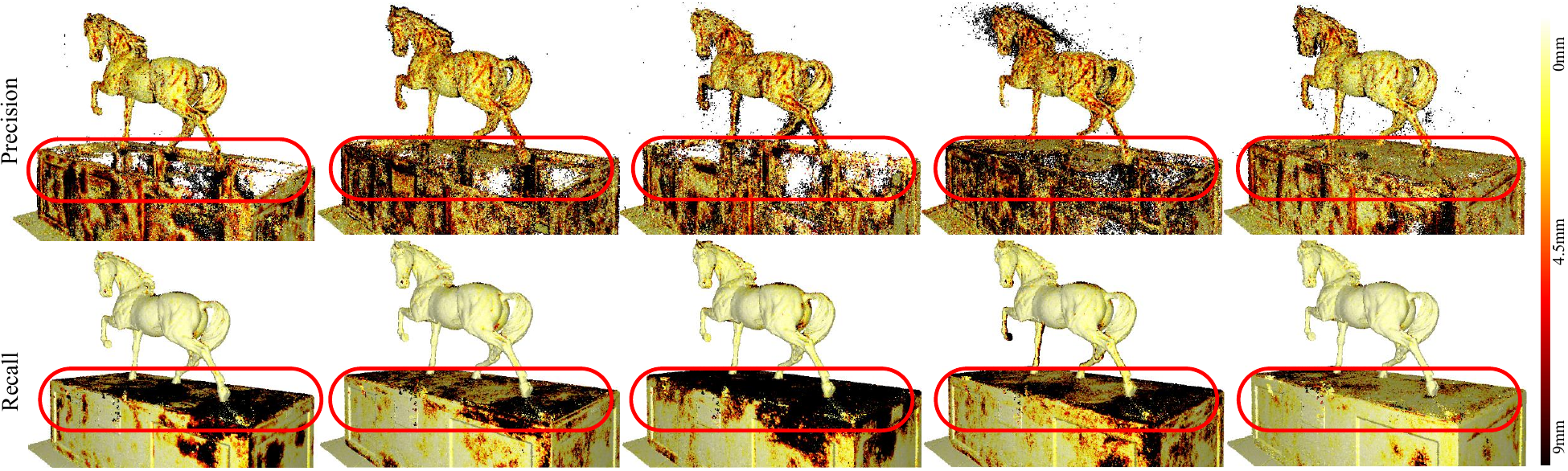}
    
\makebox[0.18\textwidth]{\scriptsize (a) MVSFormer~\cite{cao2022mvsformer}}
\makebox[0.18\textwidth]{\scriptsize (b) RA-MVSNet~\cite{zhang2023multi}}
\makebox[0.18\textwidth]{\scriptsize (c) ET-MVSNet~\cite{liu2023epipolar}}
\makebox[0.18\textwidth]{\scriptsize (d) GeoMVSNet
    ~\cite{zhang2023geomvsnet}}
\makebox[0.18\textwidth]{\scriptsize (e) Ours}
    \label{fig:head}
    \vspace{-3pt}
    \captionof{figure}{\textbf{Comparison of reconstruction errors on Tanks and Temple benchmark.} We show precision and recall error maps for the ``Horse'' scan. Our method demonstrates notable improvements over existing methods in challenging areas.}
    \vspace{-5pt}
\end{center}
}]

\begin{abstract}
Matching cost aggregation plays a fundamental role in learning-based multi-view stereo networks. However, directly aggregating adjacent costs can lead to suboptimal results due to local geometric inconsistency. 
Related methods either seek selective aggregation or improve aggregated depth in the 2D space, both are unable to handle geometric inconsistency in the cost volume effectively. 
In this paper, we propose GoMVS to aggregate geometrically consistent costs, yielding better utilization of adjacent geometries.
More specifically, we correspond and propagate adjacent costs to the reference pixel by leveraging the local geometric smoothness in conjunction with surface normals. 
We achieve this by the geometric consistent propagation (GCP) module. It computes the correspondence from the adjacent depth hypothesis space to the reference depth space using surface normals, then uses the correspondence to propagate adjacent costs to the reference geometry, followed by a convolution for aggregation. 
Our method achieves new state-of-the-art performance on DTU, Tanks \& Temple, and ETH3D datasets. Notably, our method ranks 1st on the Tanks \& Temple Advanced benchmark.
Code is available at \href{https://github.com/Wuuu3511/GoMVS}{https://github.com/Wuuu3511/GoMVS}.
\end{abstract}
\blfootnote{\textsuperscript{*} indicates equal contributions and \textsuperscript{\corr} indicates corresponding authors.}
    
\section{Introduction}
\label{sec:intro}

Multi-view stereo (MVS) is a fundamental computer vision problem that recovers 3D shapes from posed images by multi-view correspondence matching \cite{schonberger2016pixelwise}.
Recent learning-based MVS \cite{yao2018mvsnet,wang2021patchmatchnet, li2023learning, wu2024boosting} estimates scene depth from the cost volume computed by geometric matching, which delivers latent geometric cues crucial for the final depth \cite{gu2020cascade}. However, the initial cost volume can suffer from challenging matching conditions, \eg, varying illumination, textless areas, or repetitive patterns, leading to suboptimal pixel-wise costs that hamper accurate estimations. 
\par
To mitigate this issue, cost aggregation plays an important role in removing matching ambiguities and improving discriminativeness by using the neighboring information.
However, the adjacent costs may deliver \textit{inconsistent} depth cues due to the gradual changes in local geometry. As a result, the aggregated costs are not geometrically guaranteed to have the highest matching score at the real reference depth, leading to suboptimal depth predictions.
The widely adopted cascade framework \cite{gu2020cascade} can potentially exacerbate this issue as the adjacent costs can have more divergent costs due to the shifted depth hypotheses.
\par
As the geometric inconsistency is a common challenge in multi-view stereo and 2-view stereo matching, related methods either adopt learned aggregation \cite{wang2021patchmatchnet, xu2020aanet} or enforce consistency to the aggregated depth \cite{kusupati2020normal, long2020occlusion, yin2019enforcing}. Specifically, some methods \cite{wang2021patchmatchnet, xu2020aanet} adopt adaptive aggregation schemes to allow networks to select pixels that potentially correlate well and contribute to the reference pixel's geometry. However, they heavily rely on network capabilities and do not guarantee geometric plausibility from the selected costs. Other methods \cite{kusupati2020normal, qi2018geonet} seek to refine or regularize the aggregated depth values using jointly estimated surface normals.
However, these methods only refine the output depth in 2D image space and are inherently unable to handle inconsistencies in the cost volume, which is vital for MVS methods.

\par
In this paper, we propose GoMVS that aggregates geometrically consistent costs, allowing better utilization of adjacent geometries.
Considering that the local geometry is usually smooth and exhibits gradual changes, we leverage the local smoothness to correspond and propagate adjacent costs to the reference cost.
We achieve this by the geometrically consistent aggregation scheme, which operates on the local convolution window and propagates adjacent costs with the geometrically consistent propagation (GCP) module.
The GCP module computes the correspondences from the adjacent cost's hypothesized depth space to the reference cost's depth space, using back-projected depth hypotheses and the surface normal.
Then, it propagates the adjacent costs to the reference by interpolating cost scores with respect to the correspondence. After propagating adjacent costs within a local window, we aggregate them using standard convolutions. 
Unlike previous methods \cite{qi2018geonet,kusupati2020normal,long2020occlusion} that refine the predicted depth in the 2D space, our method incorporates geometric consistency in the cost space, yielding a better utilization of adjacent geometries.
As surface normal is crucial for corresponding and propagating local costs, we further investigate different choices of normal predictions.
We find appropriately applying off-the-shelf monocular normal models enables smooth and robust aggregation across datasets.
We conduct extensive experiments to evaluate our method's effectiveness, and our method achieves new state-of-the-art on DTU, Tank \& Temple, as well as the ETH3D dataset. Our contributions are summarized as follows:
\begin{itemize}[topsep=-0.0cm, itemsep=-0cm]
\item We propose GoMVS to aggregate geometrically consistent costs, allowing better utilization of adjacent geometries.
\item We propose a geometrically consistent propagation (GCP) module that allows geometrically plausible correspondence and propagation in cost space.
\item We investigate different choices of normal computation and find that properly applying the monocular surface normal model performs well across datasets. 
\end{itemize}
\section{Related Works}
\label{sec:relatedworks}
\subsection{Learning-based MVS Methods}
Multi-View Stereo (MVS) aims to reconstruct 3D scenes from multiple posed images. 
In recent years, learning-based methods have exhibited promising results.
MVSNet~\cite{yao2018mvsnet} uses differentiable homography to construct the cost volume and employs a 3D U-Net for regularization.
Subsequent works improve this framework in several ways.
RNN-based methods~\cite{yao2019recurrent,Wei_2021_ICCV,yan2020dense}and coarse-to-fine approaches~\cite{gu2020cascade,cheng2020deep,yang2020cost,mi2022generalized,wang2021patchmatchnet} reduce memory consumption through by designing efficient structures.
Another group of methods~\cite{ding2022transmvsnet, liao2022wt, liu2023epipolar} devises local or global attention modules to enhance input feature representations. 
MVSFormer~\cite{cao2022mvsformer} incorporates an additional pre-trained transformer network, enhancing the performance of MVS with a powerful feature extractor. However, it lacks further exploration in terms of geometry.
GeoMVSNet~\cite{zhang2023geomvsnet} utilizes the coarse depth map to extract additional geometric features.
In addition, ~\cite{xu2022learning,zhang2023vis,wang2022mvster} have designed pixel-wise visibility modules to handle occlusions. 
\subsection{Cost Volume Aggregation}
As cost volume is vital for multi-view depth estimation, recent works introduce different cost aggregation methods to the depth network.
NP-CVP-MVSNet~\cite{yang2022non} introduces sparse convolution to aggregate matching costs at the same depth range.
WT-MVSNet~\cite{liao2022wt} employs a cost transformer to generate a more complete and smoother probability volume.
GeoMVSNet~\cite{zhang2023geomvsnet} incorporates the coarse probability volume to enhance the matching discriminative ability.
While these methods improve the capability of regularization networks, the local geometric inconsistency of the cost volume still remains and poses challenges for the final aggregation results.
\subsection{Normal Assisted Depth Estimation}
Surface normal provides rich geometric details and has been widely applied in recent years to depth estimation tasks.
Traditional MVS methods~\cite{xu2019multi, xu2020planar, Ren_2023_ICCV} optimize depth and normal hypotheses simultaneously by constructing a planar prior model.
Inspired by traditional methods, SP-Net~\cite{wang2022spnet} performs slanted plane cost aggregation by learning parameterized local planes.
NAPV-MVS~\cite{tong2022normal} uses local normal similarity to emphasize the most relevant adjacent costs.
NR-MVSNet~\cite{li2023nr} utilizes depth-normal consistency to adaptively expand the hypothesis range, providing broader matchings to assist depth inference.
However, these methods do not address the local inconsistency issue.
GeoNet~\cite{qi2018geonet} proposes a monocular depth estimation method that uses kernel regression to refine output depth with normals.
However, it is sensitive to noisy outputs and is inherently incapable of handling cost volume inputs.
Another line of works ~\cite{kusupati2020normal, long2020occlusion, yin2019enforcing} proposes the depth-normal consistency loss to enhance the network's perception of geometric cues.
Unlike these methods, our method leverages the normal to yield geometrically consistent costs in the 3D space, yielding better utilization of adjacent costs.
\begin{figure*}[t]
\begin{center}
\includegraphics[width=0.95\linewidth]{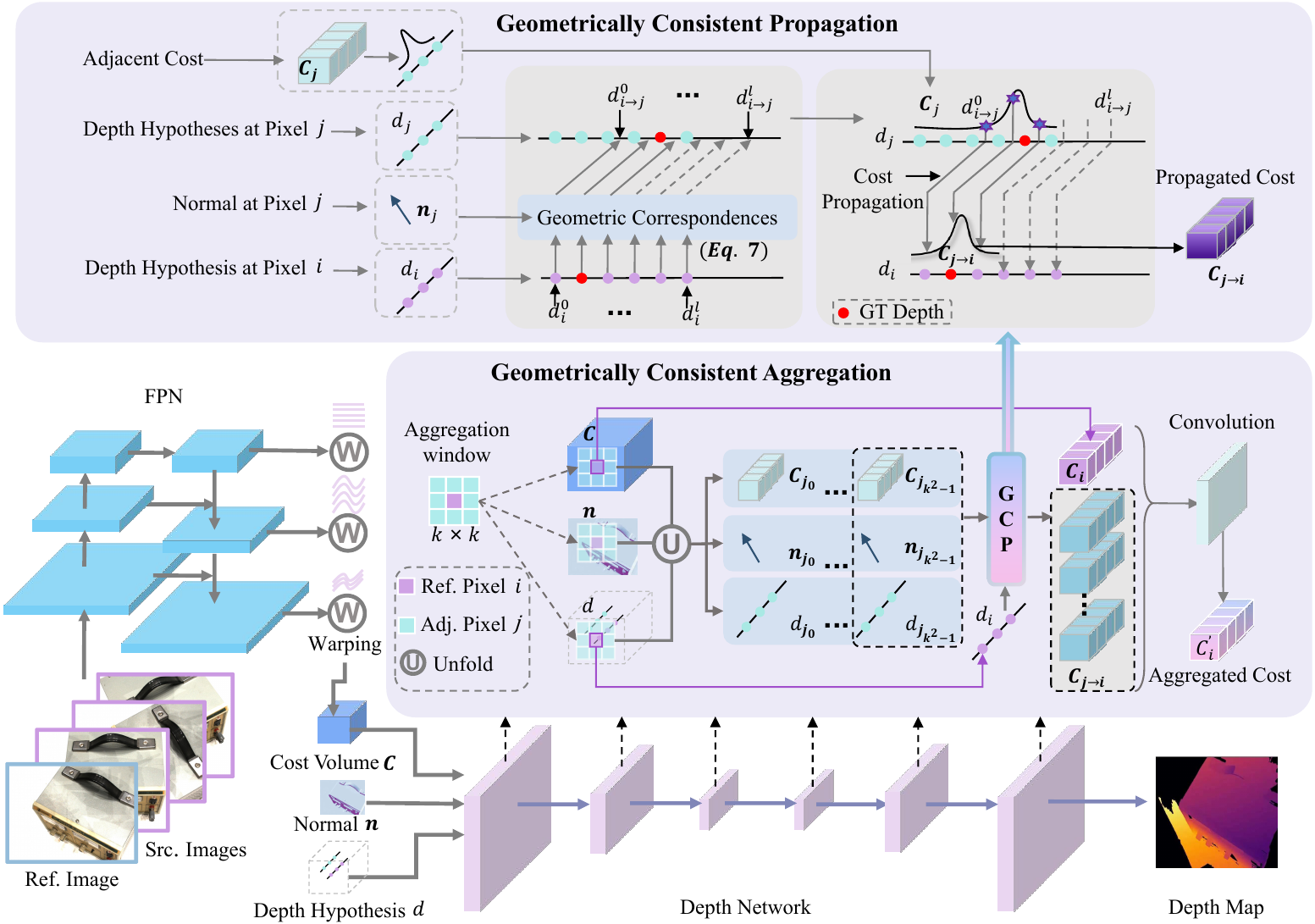}
\end{center}
\vspace{-15pt}
\caption{\textbf{Overview of our method.} Given a reference image and a set of source images, we use FPN to extract multi-scale features for cost volume reconstruction. To conduct geometrically consistent aggregation within the local window, we collect adjacent geometric cues and send them to the proposed geometrically consistent propagation (GCP) module, which computes the correspondence from the adjacent depth hypothesis space to the reference depth space. The resulting costs are endowed with geometric consistency, which facilitates better utilization of adjacent geometry and can be aggregated by the convolution.}
\label{fig:networkarc}
\end{figure*}

\vspace{-5pt}
\section{Methodology}
\label{sec:methodology}
Given a reference image $\mathbf{I}_0 \in \mathbb{R}^{H\times W \times 3}$ and $N$ source images $\{\mathbf{I}_{i}\}{_{i=1}^{N}}$, as well as camera intrinsic $\{\mathbf{K}_{i}\}_{i=0}^{N}$ and extrinsic parameters $\{[\mathbf{R}_{0\rightarrow i};\mathbf{t}_{0 \rightarrow i}]\}_{i=1}^{N}$, our goal is to estimate the depth map of $\mathbf{I}_0$ from multiple posed images.
Fig.~\ref{fig:networkarc} shows an overview of our method.
We first utilize multi-scale image features to build the cost volume (Sec.~\ref{Volume_Construction}).
We then introduce the geometrically consistent aggregation scheme (Sec.~\ref{cost_reg}), which consists of the blocks in the depth network. We then investigate different choices for obtaining surface normals (Sec. \ref{get_normal_cues}). 
\subsection{Cost Volume Construction}\label{Volume_Construction}

We first apply a Feature Pyramid Network~\cite{lin2017feature} to extract multi-scale image features $\{\mathbf{F}_i^s\}_{i=0}^{N} \in \mathbb{R}^{\frac{H}{2^s}\times \frac{W}{2^s} \times M}$, where $s$ is the scale factor.
For simplicity, we omit the superscript of $s$ below.
To build the cost volume in each stage, we first 
sample depth hypotheses $d$ for each pixel in a predefined depth range.
Through differentiable homography, we can compute the corresponding position $\mathbf{p}'$ of the reference image's pixel $\mathbf{p}$ in the source image,

\begin{equation}
\label{homography}
\mathbf{p}'=\mathbf{K}_{i} [ \mathbf{R}_{0\rightarrow i}(\mathbf{K}_{0}^{-1}\mathbf{p} d)+\mathbf{t}_{0\rightarrow i}],
\end{equation}
where $\mathbf{R}$ and $\mathbf{t}$ denote the rotation and translation parameters and 
 $\mathbf{K}$ are the intrinsic matrix.
Let ${\mathbf{F}(\mathbf{p})}$ represents the feature vector at pixel $\mathbf{p}$, then the two-view feature correlation volume $\mathbf{V}$ at pixel $\mathbf{p}$ can be represented as
\begin{equation}
\label{two_view_cost}
\mathbf{V}_{i}(\mathbf{p})=\mathbf{F}_{0}(\mathbf{p}) \cdot \mathbf{F}_{i}({\mathbf{p}'}),
\end{equation}
where $\cdot$ refers to the dot product. To aggregate multiple pair-wise cost volumes, we utilize a shallow network~\cite{wang2021patchmatchnet} to learn the pixel-wise weight maps $\mathbf{W}$. The weight computation takes place exclusively in the initial stage, while weight maps for subsequent stages are derived through upsampling from the previous stage.
Then the multi-view aggregated cost volume $\mathbf{C}$ can be represented as:
\begin{equation}
\label{two_view_cost}
\mathbf{C}=\frac{{\sum_{i=1}^N}{\mathbf{W}_i \odot \mathbf{V}_i}}{{\sum_{i=1}^N}{\mathbf{W}_i}}.
\end{equation}
\subsection{Geometrically Consistent Aggregation}\label{cost_reg}
An essential idea of cost aggregation is to leverage neighboring information to improve the discriminativeness of the cost volume, where the key is to find the most relevant neighbors and effectively aggregate their matching costs. To achieve this, typical convolution-based methods are limited to the size of the convolution kernel (\eg, $3 \times 3 \times 3$), and geometric inconsistency is very likely to happen in this local region due to non-constant depth distributions within this kernel. It's also computationally inefficient to directly increase the kernel size to get improved performance. 

In this paper, we observe in a small local region, many scenes can be approximated with a plane, which frequently exists in real-world scenarios. To this end, we propose to leverage this locally approximated planar structure to guide the cost aggregation process in a geometrically consistent manner. There exists an analytic relationship between the reference pixel's depth and its local neighbors, which can be leveraged to obtain more reliable cost candidates.
Specifically, for each reference pixel, we first collect the geometric clues of its $k\times k$ spatial window to compute the correspondences of the depth hypothesis. 
Depending on the corresponding location, we propagate the adjacent costs to the reference pixel's depth space.
Finally, we use a convolution layer to aggregate the propagated costs.

\vspace{-10pt}
\subsubsection{Local Geometric Clues Collection}\label{Decomposing the local cost}
We first collect local depth hypotheses and normal maps for each pixel within a spatial window.
Specifically, given the depth hypotheses of shape $L\times H\times W$ and the normal map of shape $ 3\times H\times W$, where $L$ is the depth hypothesis number and $H$, $W$ denotes the spatial dimension, we unfold each pixel with a $k\times k$ spatial window, yielding local intermediate depth hypotheses volume and normal map of shape $ k^2\times L\times H\times W$ and $ k^2\times 3\times H\times W$, respectively. We then compute the depth hypothesis correspondences based on these intermediate geometric clues.

\subsubsection{Geometrically Consistent Propagation}\label{Geometric consistency propagation}

To better aggregate the high-quality costs of the adjacent pixels, we align the adjacent pixels' depth hypothesis to the depth space of the reference pixel.
Based on depth correspondence, we perform geometrically consistent cost propagation (GCP).
Firstly, we introduce the depth relationship among pixels within the same plane.
Given a pixel's image coordinates $(u, v)$ and depth $d(u, v)$, its 3D point $X(u, v)$ in the camera coordinate system can be represented as
\begin{equation}
\label{cal3dpoint}
X(u,v)= 
\left[
 \begin{array}{ccc}
     x \\
     y \\
     z
 \end{array}
 \right]
 =
\left[
 \begin{array}{ccc}
     \frac{u-c_x}{f_x} \\
     \frac{v-c_y}{f_y} \\
     {1}
 \end{array}
 \right]
 {d(u, v)}
,
\end{equation}
where $c_x$, $c_y$, $f_x$, and $f_y$ are the parameters of camera intrinsic $\mathbf{K}$.
For the given reference pixel $i$ and adjacent pixel $j$,
we model the relationship between $X(u_i,v_i)$ and $ X(u_j,v_j)$ by leveraging local planar assumption and the surface normal $\mathbf{n}$. They satisfy the equation of
\begin{equation}
\label{two_point_plane_equation}
{\mathbf{n}^\top}(X(u_i,v_i) - X(u_j,v_j)) = 0.
\end{equation}
According to Eq.~\eqref{cal3dpoint} and Eq.~\eqref{two_point_plane_equation}
, the depth relationship between the reference pixel $i$ and the adjacent pixels $j$ can be represented as:
\begin{equation}
\label{depthj/depthi}
\frac{d(u_j,v_j)}{d(u_i,v_i)}
=
\frac{{\mathbf{n}^\top}\left[
 \begin{array}{ccc}
     \frac{u_i-c_x}{f_x} & \frac{v_i-c_y}{f_y} & {1}
 \end{array}
 \right]^\top}{{\mathbf{n}^\top}\left[
 \begin{array}{ccc}
     \frac{u_j-c_x}{f_x} & \frac{v_j-c_y}{f_y} & {1}
 \end{array}
 \right]^\top}
.
\end{equation}
We use
 $r_{ji} = \frac{d(u_j,v_j)}{d(u_i,v_i)}$ to denote the depth ratio between $j$ and $i$, which describes the linear transformation of depth within the plane.
Based on this, we can compute the depth hypothesis correspondences.
Specifically, define $[d^{1}_{i},...,d^{L}_{i}]$ as the depth hypothesis in the pixel $i$'s depth space,  where $L$ refers to the number of depth sampling levels.
Each depth hypothesis is then mapped to pixel $j$'s depth space through the depth ratio $r_{ji}$.
\begin{equation}
\label{di->j}
[d^{1}_{i\rightarrow j},..., d^{L}_{i\rightarrow j}]
=
[r_{ji} \times d^{1}_{i},...,r_{ji} \times d^{L}_{i}]
,
\end{equation}
where $d_{i\rightarrow j}$ represents the mapping depth of pixel $i$'s depth hypothesis in pixel $j$'s depth space.
We then propagate the matching cost of pixel $j$ at the $d_{i\rightarrow j}$ to $d_{i}$.
Let $\mathbf{C}_j$ denote the cost for pixel $j$. 
The propagated matching cost $\mathbf{C}_{j\rightarrow i}$ can be expressed as:
\begin{equation}
\label{C_jiprop}
\mathbf{C}_{j\rightarrow i}(d^{0}_{i},...,d^{l}_{i})
=
\mathbf{C}_{j}(d^{0}_{i\rightarrow j},..., d^{l}_{i\rightarrow j})
.
\end{equation}
Since depth hypotheses are discretely sampled at regular depth intervals within the depth range, we can
conveniently use linear interpolation to implement the above process.
With the definition $d^{m}_{i\rightarrow j}=d^{n}_{j}$, $\mathbf{C}_{j\rightarrow i}(d^{m}_{i})$ can be expressed as:
\begin{equation}
\label{linaner}
\mathbf{C}_{j\rightarrow i}(d^{m}_{i})
=
(\mathbf{C}_{j}(d^{\lceil n \rceil}_{j}) - \mathbf{C}_{j}(d^{\lfloor n \rfloor}_{j}))\frac{n - \lfloor n \rfloor}{\lceil n \rceil - \lfloor n \rfloor}
.
\end{equation}
We refer to this process as geometrically consistent propagation from $j$ to $i$. 
It can generate geometrically consistent cost candidates for each reference pixel.
Due to varying depth relationships between each pixel and its adjacent pixels, cost propagation generates an intermediate cost of $k^2M\times L\times H\times W$, where $M$ is the channel dimension.

\vspace{-5pt}
\subsubsection{Aggregating Propagated Costs}\label{Propagated cost aggregation}
Since the intermediate costs include $k\times k$ spatial information in the channel dimension, we thus aggregate the costs using convolutions with a kernel size of $1\times 1 \times k$ and an expanded channel dimension $k^2M$, leading to the same parameters as the generic 3D convolutions with kernel size $k\times k\times k$.

We encapsulate GCP and the convolution into one geometrically consistent aggregation operator used to build the depth network.
In particular, we still keep the 3D U-Net architecture proposed by MVSNet~\cite{yao2018mvsnet}, while replacing each standard 3D convolution block with our proposed geometrically consistent aggregation operator.
For the upsampling layer in the U-Net structure, we use the pixel shuffle to reorganize features and obtain a high-resolution cost volume.

\subsection{Extracting Normal Cues}\label{get_normal_cues}

Since our approach uses the surface normal for cost aggregation, in this section, we study different methods for obtaining surface normals. We conduct experiments to demonstrate the effectiveness of each method in Sec. \ref{sec:exp_abla}.

\noindent \textbf{Depth to normal.}
The surface normal can be directly computed from the estimated depth. Since we use a three-stage cascade structure, we leverage the depth map from the $g$ stage to generate the surface normal for the $g+1$ stage. 
The normal $\mathbf{n}$ can be computed \cite{qi2018geonet} in closed form as:
\begin{equation}
\label{depth2normal}
\mathbf{n}
=
\frac
{(\mathbf{A}^\mathsf{T}\mathbf{A})^{-1}{\mathbf{A}^\mathsf{T}}{\mathbf{1}}}
{\Vert {(\mathbf{A}^\mathsf{T}\mathbf{A})^{-1}{\mathbf{A}^\mathsf{T}}{\mathbf{1}}} \Vert},
\end{equation}
where $\mathbf{A}$ is a matrix composed of the coordinates of all pixels within the local window. In addition to using estimated depth maps, we also compute the GT normal from the GT depth maps following the same protocol and use it to train our method for evaluating performances.

\noindent \textbf{Cost to normal.}
In addition,
inspired by ~\cite{kusupati2020normal}, we use an additional network branch to directly regress the normal map from the cost volume in each stage, which is then used as a prior for geometrically consistent aggregation.

\noindent \textbf{Off-the-shelf monocular surface normal.}
Monocular networks directly perceive surface geometry from deep features and can estimate reasonable solutions in regions with multi-view consistency ambiguities, which complements the task of MVS.
Therefore, we explore an existing monocular normal estimation network Omnidata~\cite{eftekhar2021omnidata} to generate the surface normal.
Since Omnidata is trained on low-resolution images, its normal prediction might become unreliable when the testing input resolution is increased. To tackle this, we adopt a divide-and-conquer approach following MonoSDF~\cite{yu2022monosdf} to generate high-resolution normal cues. 
Specifically, we first divide the high-resolution image into multiple overlapping patches. Surface normal estimation is then independently conducted for each patch. Subsequently, the surface normal results are aligned and fused to generate a high-resolution normal map.

\subsection{Optimization}\label{lf}
We treat the MVS task as a classification problem and employ the winner-takes-all strategy to obtain the final depth map \cite{yao2019recurrent}.
We use the cross-entropy loss (Eq. \ref{loss}) in each stage, which is applied to the probability volume $P$ and the ground truth one-hot volume $P^{'}$.
Following ~\cite{mi2022generalized}, all depth out-of-range will be masked during the training stage.
\begin{equation}
\mathcal{L}=\sum_{i=1}^{d}{-}P^{'}_{i}\log(P_{i}).
\label{loss}
\end{equation}

\begin{figure*}[h]
\begin{center}
\includegraphics[scale=0.55]{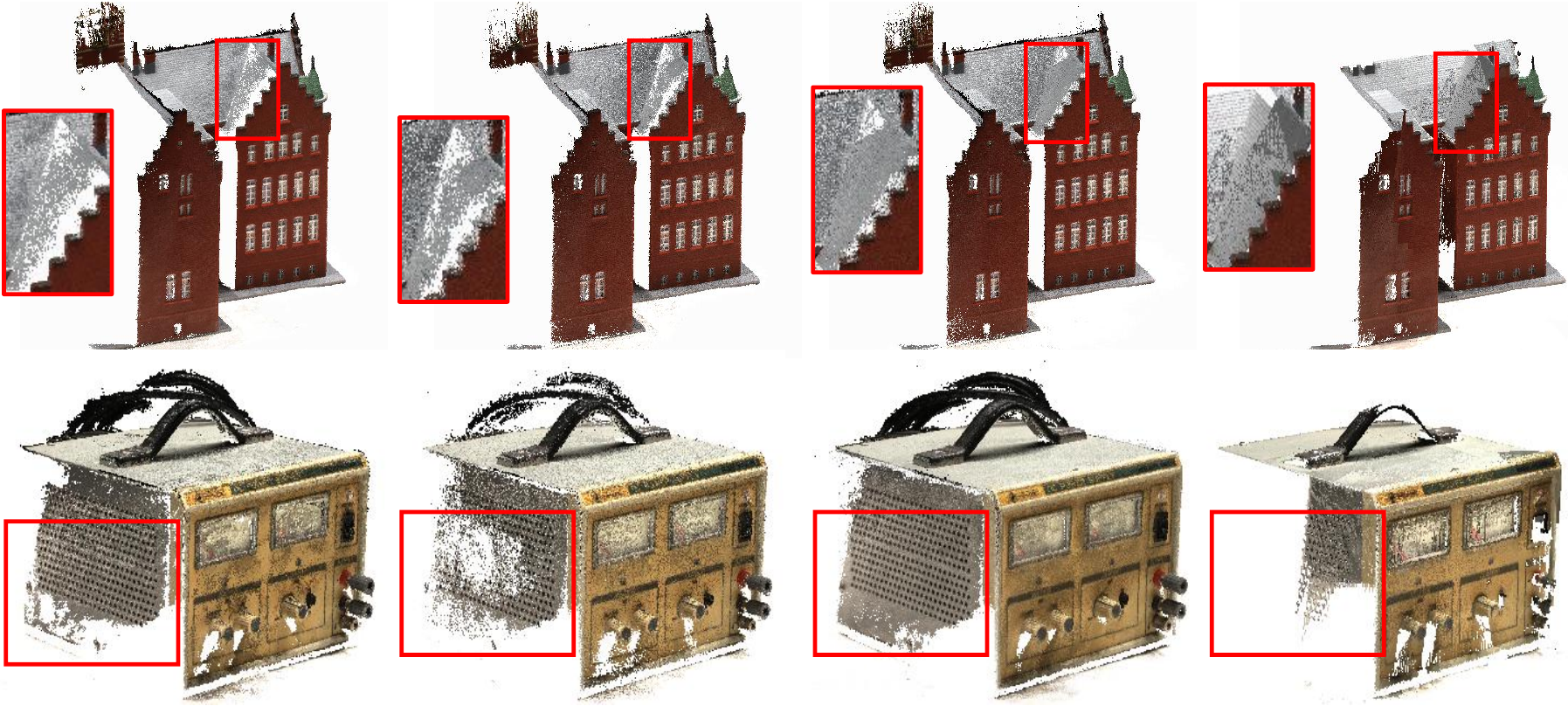}%

\makebox[0.24\textwidth]{\scriptsize (a) TransMVSNet \cite{ding2022transmvsnet}}
\makebox[0.24\textwidth]{\scriptsize (b) GeoMVSNet \cite{zhang2023geomvsnet}}
\makebox[0.24\textwidth]{\scriptsize (c) \textbf{Ours}}
\makebox[0.24\textwidth]{\scriptsize (d) GT}
\end{center}
    \vspace{-10pt}
   \caption{\textbf{Comparison of reconstruction results.} 
   Our method reconstructs more complete results in challenging areas.}
\vspace{-10pt}
\label{fig:dtu11and29}
\end{figure*}

\vspace{-10pt}
\section{Experiments}
\label{sec:experiments}
In this section, we evaluate our method on the DTU~\cite{aanaes2016large}, ETH3D~\cite{schoeps2017cvpr}, and Tanks and Temple~\cite{knapitsch2017tanks} datasets, respectively.
Furthermore, we conducted multiple ablation experiments on the DTU dataset to validate the effectiveness of our method.
\subsection{Datasets}
DTU~\cite{aanaes2016large} dataset comprises 128 scenes in controlled laboratory environments, with models captured using structured light scanners. 
Each scene was scanned from the same 49 or 64 camera positions under 7 different lighting conditions.
The official evaluation assesses the point cloud using distance metrics of accuracy and completeness.
BlendedMVS~\cite{yao2020blendedmvs} is a large-scale MVS dataset that consists of over 17,000 high-resolution images covering a variety of scenes, including urban environments, architecture, sculptures, and small objects. 
Tanks and Temples (TNT)~\cite{knapitsch2017tanks} is a real-world dataset, divided into two sets, including 8 scenes in the intermediate set and 6 scenes in the advanced set.
ETH3D~\cite{schoeps2017cvpr} dataset consists of multiple indoor and outdoor scenes with large viewpoint variations.
The quality of point clouds on the ETH3D and TNT datasets is measured using the percentage of precision and recall.
\subsection{Implementation Details}
\paragraph{Training}
Following the data partitioning of MVSNet, we first train the model on the DTU training set. 
Our network employs a three-stage cascade structure, with depth sampling at 48, 32, and 8 in each stage and depth intervals of 4, 1, and 0.5, respectively. 
We train our model with 5 input images, each having a resolution of 512$\times$640. 
The model is optimized using Adam for 12 epochs, starting with an initial learning rate of 0.001 which is reduced by 0.5 after the 6 and 8 epochs.
We then fine-tune the model on the BlendedMVS dataset with 9 images at a resolution of 576$\times$768 for evaluation on Tanks and Temples and ETH3D datasets.
During fine-tuning, we reduce the depth sampling interval of the last stage by half of its original value.

\vspace{-10pt}
\paragraph{Evaluation}
When testing on the DTU dataset, we use 5 images at a resolution of 864$\times$1152 as input and employ the depth map filtering method following~\cite{zhang2023geomvsnet} to generate the final point cloud. 
For the tanks and temple dataset, we carried out tests using 11 images with a resolution of 960$\times$1920. 
In terms of depth map fusion, we employ the widely adopted dynamic fusion strategy~\cite{yan2020dense}.
Moreover, we conducted tests on the ETH3D dataset using images with a size of 1152$\times$1536 and the depth map fusion strategy is consistent with IterMVS~\cite{wang2022itermvs}.
\begin{figure*}[h]
\begin{center}
\includegraphics[scale=0.54]{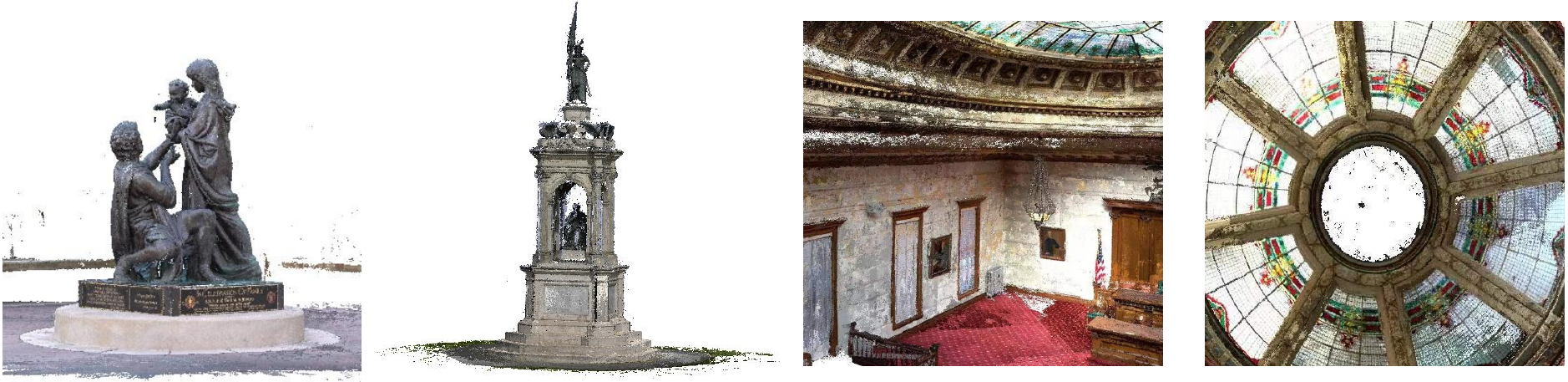}%

\makebox[0.24\textwidth]{\scriptsize (a) Family}
\makebox[0.24\textwidth]{\scriptsize (b) 
Francis}
\makebox[0.24\textwidth]{\scriptsize (c) Courtroom}
\makebox[0.24\textwidth]{\scriptsize (d) Musume}
\end{center}
    \vspace{-10pt}
   \caption{\textbf{Qualitative results on Tanks and Temples.} Our method achieves detailed and complete reconstructions across different scenes.}
\label{fig:tnt_pre_recall}
\end{figure*}

\subsection{Benchmark Performance}
\paragraph{Evaluation on DTU dataset.}
\begin{table}
 \renewcommand{\arraystretch}{1}
 \setlength{\tabcolsep}{15pt}
 \centering
 \scalebox{0.78}{
\begin{tabular}{lccc}
\toprule
Method      & Acc. $\downarrow$& Comp. $\downarrow$& Overall$\downarrow$ \\ 
\midrule
Gipuma~\cite{galliani2015massively}      & \textbf{0.283}    & 0.873     & 0.578       \\
COLMAP~\cite{schonberger2016pixelwise}      & 0.400    & 0.664     & 0.532       \\
NAPV-MVS~\cite{tong2022normal} &0.367 &0.375 &0.371 \\ 
AA-RMVSNet~\cite{Wei_2021_ICCV}  & 0.376    & 0.339     & 0.357       \\
Vis-MVSNet~\cite{zhang2023vis}  & 0.369    & 0.361     & 0.365       \\
CasMVSNet~\cite{gu2020cascade}   & 0.325    & 0.385     & 0.355       \\
UniMVSNet~\cite{peng2022rethinking}   & 0.352    & 0.278     & 0.315       \\
MVSTER~\cite{wang2022mvster}      & 0.350    & 0.276     & 0.313       \\
TransMVSNet~\cite{ding2022transmvsnet} & 0.321    & 0.289     & 0.305       \\
GbiNet*~\cite{mi2022generalized}   & 0.314    & 0.295     & 0.305            \\ 
RA-MVSNet~\cite{zhang2023multi}   & 0.326    & 0.268     & 0.297  \\
GeoMVSNet~\cite{zhang2023geomvsnet}   & 0.331    & 0.259     & 0.295            \\ 
ET-MVSNet~\cite{liu2023epipolar}   & 0.329    & 0.253     & 0.291            \\ 

MVSformer~\cite{cao2022mvsformer}   & 0.327    & 0.251     & 0.289           \\ 
\midrule
\textbf{GoMVS}         & 0.347    & \textbf{0.227}     & \textbf{0.287}       \\ 
\bottomrule \\
\end{tabular}}
\vspace{-10pt}
\caption{\textbf{Quantitative results on DTU~\cite{aanaes2016large}}.
Our method achieves the best completeness and overall score. Moreover, the completeness of our point cloud outperforms previous methods by large margins. 
}
\vspace{-15pt}
\label{table:dtu_result}
\end{table}

We compare both traditional methods and deep learning-based approaches.
The quantitative evaluation results for point cloud reconstruction are shown in Tab \ref{table:dtu_result}. 
Our method achieves SOTA completeness and overall performance.
It is worth noting that our method shows obvious improvement in completeness compared to previous methods. 
This demonstrates that our method can better use adjacent costs to propagate local geometries, resulting in a more complete reconstruction.
Fig. \ref{fig:dtu11and29} shows a comparison of our point cloud results with previous SOTA methods. 
We have more detailed and complete reconstructions in the challenge areas.

\begin{table*}[]
 \centering
 \renewcommand{\arraystretch}{1.2}
 \setlength{\tabcolsep}{8pt}
 \scalebox{0.68}{
\begin{tabular}{lcccccccccccccccc}
\toprule
\multirow{2}{*}[-0.5ex]{Methods} & \multicolumn{9}{c}{Intermediate}                                     & \multicolumn{7}{c}{Advanced}                          \\ 
\cmidrule(lr){2-10} \cmidrule(lr){11-17}
                         & Mean$\uparrow$  & Fam.  & Fra.  & Hor.  & Lig.  & M60   & Pan.  & Pla.  & Tra.  & Mean$\uparrow$  & Aud.  & Bal.  & Cou.  & Mus.  & Pal.  & Tem.  \\ 
                         \midrule
COLMAP~\cite{schonberger2016pixelwise}                   & 42.14 & 50.41 & 22.25 & 26.63 & 56.43 & 44.83 & 46.97 & 48.53 & 42.04 & 27.24 & 16.02 & 25.23 & 34.70 & 41.51 & 18.05 & 27.94 \\
CasMVSNet~\cite{gu2020cascade}                & 56.84 & 76.37 & 58.45 & 46.26 & 55.81 & 56.11 & 54.06 & 58.18 & 49.51 & 31.12 & 19.81 & 38.46 & 29.10 & 43.87 & 27.36 & 28.11 \\
Vis-MVSNet~\cite{zhang2023vis}               & 60.03 & 77.40 & 60.23 & 47.07 & 63.44 & 62.21 & 57.28 & 60.54 & 52.07 & 33.78 & 20.79 & 38.77 & 32.45 & 44.20 & 28.73 & 37.70 \\
GBiNet~\cite{mi2022generalized}                  & 61.42 & 79.77 & 67.69 & 51.81 & 61.25 & 60.37 & 55.87 & 60.67 & 53.89 & 37.32 & 29.77 & 42.12 & 36.30 & 47.69 & 31.11 & 36.93 \\
EPP-MVSNet~\cite{ma2021epp}               & 61.68 & 77.86 & 60.54 & 52.96 & 62.33 & 61.69 & 60.34 & 62.44 & 55.30 & 35.72 & 21.28 & 39.74 & 35.34 & 49.21 & 30.00 & 38.75 \\
TransMVSNet~\cite{ding2022transmvsnet}              & 63.52 & 80.92 & 65.83 & 56.94 & 62.54 & 63.06 & 60.00 & 60.20 & 58.67 & 37.00 & 24.84 & 44.59 & 34.77 & 46.49 & 34.69 & 36.62 \\

UniMVSNet~\cite{peng2022rethinking}                 & 64.36 & 81.20 & 66.43 & 53.11 & 64.36 & 66.09 & 64.84 & 62.23 & 57.53 & 38.96 & 28.33 & 44.36 & 39.74 & 52.89 & 33.80 & 34.63 \\  
D-MVSNet~\cite{Ye_2023_ICCV}                 & 64.66 & 81.27 & 67.54 & 59.10 & 63.12 & 64.64 & 64.80 & 59.83 & 56.97 & 41.17 & 30.08 & 46.10 & 40.65 & \textbf{53.53} & 35.08 & 41.60 \\ 
ET-MVSNet~\cite{liu2023epipolar}                 & 65.49 & 81.65 & 68.79 & 59.46 & 65.72 & 64.22 & 64.03 & 61.23 & 58.79 & 40.41 & 28.86 & 45.18 & 38.66 & 51.10 & 35.39 & 43.23 \\ 
RA-MVSNet~\cite{zhang2023multi}                 & 65.72 & 82.44 & 66.61 & 58.40 & 64.78 & \textbf{67.14} & 65.60 & 62.74 & 58.08 & 39.93 & 29.17 & 46.05 & 40.23 & 53.22 & 34.62 & 36.30 \\ 
GeoMVSNet~\cite{zhang2023geomvsnet}                 & 65.89 & 81.64 & 67.53 & 55.78 & 68.02 & 65.49 &\textbf{67.19} & \textbf{63.27} & 58.22 & 41.52 & 30.23 & 46.53 &39.98 & \textbf53.05 & 35.98 & 43.34 \\   
MVSFormer~\cite{cao2022mvsformer}                 & 66.37 & 82.06 & \textbf{69.34} & 60.49 & \textbf{68.61} & 65.67 &64.08 & 61.23 & 59.33 & 40.87 & 28.22 & 46.75 &39.30 & \textbf52.88 & 35.16 & 42.95 \\   
\midrule

\textbf{GoMVS}                      & \textbf{66.44}& \textbf{82.68} &69.23 & \textbf{69.19} &63.56 &65.13 & 62.10 & 58.81 & \textbf{60.80} & \textbf{43.07} & \textbf{35.52} & \textbf{47.15} &  \textbf{42.52 }& 52.08 & \textbf{36.34} & \textbf{44.82} \\ 
\bottomrule
\end{tabular}}
\vspace{-5pt}
\caption{\textbf{Quantitative results of F-score on Tanks and Temples benchmark.}
Our method achieves the best F-score on both the ``Intermediate'' and the challenging ``Advanced'' set.
Note that our method \textit{ranks 1st on the official TNT Advanced Benchmark}.}
\vspace{-10pt}
\label{table:tnt_result}
\end{table*}

\vspace{-15pt}
\paragraph{Evaluation on Tanks and Temples dataset.}
We validated the generalization of our model on the Tanks and Temples dataset, and the quantitative results are shown in Table \label{table:tnt_result}. We achieved the best performance on both the intermediate and advanced sets. Moreover, we \textit{ranked $1st$ among all submitted results on the advanced set of the TNT benchmark},  which contains more complex scenes. It demonstrates the strong robustness and generalization ability of our method.
Fig. \ref{fig:tnt_pre_recall} shows point cloud results on intermediate and advanced sets.
Our method achieves detailed and complete reconstructions across different indoor and outdoor scenes.

\begin{table}[t]
\centering
 \renewcommand{\arraystretch}{1}
 \setlength{\tabcolsep}{6pt}
\scalebox{0.8}{\begin{tabular}{lccc}
\toprule
Method                       & Acc. $\downarrow$ & Comp. $\downarrow$ & Overall $\downarrow$ \\ 
\midrule
Standard 3D-convolution~\cite{yao2018mvsnet}      &  0.365 &0.265    &0.315         \\
Sparse convolution~\cite{yang2022non}  &   0.354   & 0.268    & 0.311  \\
Spatial deformable aggregation~\cite{wang2021patchmatchnet} &   0.363   &0.257   &0.310         \\
Depth kernel regression~\cite{qi2018geonet}     &    0.369   &  0.262     &  0.316       \\ \hline
Ours (GCA)     & \textbf{0.347} & \textbf{0.227} & \textbf{0.287}         \\ 
\bottomrule
\end{tabular}}
\caption{\textbf{Comparison with different aggregation methods.} Our method significantly outperforms previous cost volume aggregation methods.}
\vspace{-17pt}
\label{table:abs_different_agg_methods}
\end{table}

\vspace{-15pt}
\paragraph{Evaluation on ETH3D dataset.}
The ETH3D dataset contains many challenging scenes, including scenes with textureless areas and large viewpoint variations. 
We compare our methods with previous methods and results are shown in Tab. \ref{table::eth3d}.
Our method achieves the best performance on both the validation set and the test split. In particular, it outperforms previous SOTA by a significant margin on the test split, demonstrating its generalization ability over existing methods.
\begin{table*}[]
 \centering
 \renewcommand{\arraystretch}{1.2}
 \setlength{\tabcolsep}{20pt}
\scalebox{0.7}{
\begin{tabular}{lcccccc}
\toprule
\multirow{2}{*}[-0.5ex]{Methods} & \multicolumn{3}{c}{Training} & \multicolumn{3}{c}{Test}      \\ 
\cmidrule(lr){2-4} \cmidrule(lr){5-7}
                         & Acc.$\uparrow$          & Comp. $\uparrow$          & F-score$\uparrow$              & Acc.$\uparrow$            & Comp.  $\uparrow$          & F-score$\uparrow$ \\ 
                         \midrule
COLMAP~\cite{schonberger2016pixelwise} & \textbf{91.85}   &55.13            &67.66                 &\textbf{91.97}   &62.98             &73.01              \\ 
ACMM~\cite{xu2019multi}                  & 90.67          &70.42            &78.86                 &90.65            &74.34             &80.78              \\ 
\midrule
IterMVS~\cite{wang2022itermvs}              & {73.62}   &61.87            &66.36                 &76.91   &72.65             &74.29              \\ 
GBi-Net~\cite{mi2022generalized}                  & 73.17         &69.21            &70.78                 &80.02           &75.65             &78.40   \\
MVSTER~\cite{wang2022mvster}                  & 76.92          &68.08            &72.06                 &77.09            &82.47             &79.01   \\

PVSNet~\cite{xu2022learning}                  & 83.00          &71.76            &76.57                &81.55            &83.97             &82.62   \\ 
EPP-MVSNet~\cite{ma2021epp}             & 82.76           &67.58            &74.00                 &85.47            &81.79             &83.40              \\ 
Vis-MVSNet~\cite{zhang2023vis}              & 83.32          &65.53   & {72.77}       &86.86            &80.92    &83.46              \\ 
EPNet~\cite{su2023efficient}              & 79.36          &\textbf{79.28}   & {79.08}       &80.37            &\textbf{87.84}    &83.72              \\ 
\hline
GoMVS   &  81.22      & 77.65  &  \textbf{79.16}      &85.50            &86.85    &\textbf{85.91}\\ 
\bottomrule
\end{tabular}
}
\caption{\textbf{Quantitative results on ETH3D dataset}.
We show comparisons of reconstructed point clouds using percentage metric (\%) at a threshold of 2cm. Our
approach achieves the best performance with notable margins.}\vspace{-11pt}
\label{table::eth3d}
\end{table*}

\subsection{Ablation Study}\label{sec:exp_abla}

\paragraph{Comparison with different aggregation methods.}
To verify the effectiveness of utilizing adjacent geometry, we compare different cost aggregation and depth aggregation methods, and the results are shown in Tab. \ref{table:abs_different_agg_methods}.
Regarding the cost aggregation methods,
Sparse convolution~\cite{yang2022non} aggregates the cost at the same depth without fully considering the depth geometry, resulting in certain improvements in performance compared with the baseline.
PatchMatchNet~\cite{wang2021patchmatchnet} utilizes deformable convolutions to gather spatial matching costs and aggregate them using a lightweight 3D CNN. 
We replace the aggregation network with a 3D U-Net to ensure a fair comparison with the same parameter scale. 
PatchmatchNet heavily relies on network capabilities and does not guarantee geometric plausibility from the selected costs. As a result, it brings limited performance improvements (row \#3).

Additionally, for the depth aggregation method, we refine the depth map on the baseline method by incorporating depth kernel regression proposed by GeoNet~\cite{qi2018geonet}. 
Using normal similarity to compute depth aggregation weights is prone to the influence of normal noise and cannot effectively utilize the abundant geometric information in the cost volume. 
This leads to a decline in the accuracy of the final point cloud (row \#4).
We utilize normal priors to guide cost aggregation, alleviating the challenge of geometric inconsistency and achieving the best performance among all aggregation methods.

\vspace{-5pt}
\paragraph{Comparison with different depth receptive fields.}
Intuitively, 3D convolutions with larger receptive fields in the depth dimension can alleviate the cost inconsistency in the local range, by resorting to wider areas.
Therefore, we compare our approach with variants directly expanding the depth receptive field.
We keep the $3\times3$ spatial window size at each 3D convolution layer and experiment with kernel sizes of 3, 5, and 7 in the depth dimension on the baseline method.
The quantitative results are shown in Tab. \ref{table::dif_ganshouye}, we find that increasing the receptive field in the depth dimension leads to some certain improvement. 
However, due to the lack of geometric awareness, its performance is saturated when the dimension expands to a certain kernel size. 
In contrast, we use surface normal to geometrically guide the cost aggregation process. 
With a kernel size of only 3, our method achieves the best performance, outperforming other alternatives by clear margins.

\vspace{-15pt}
\paragraph{Evaluation of different normal cues.}
Since the surface normal is important for guiding geometrically consistent aggregation, we further evaluate the effectiveness of different normal cues in Tab. \ref{table::dif_normal}. 
We first train and evaluate our method using the GT normal, which sets an upper bound for our method. As shown in the last row, it significantly improves the performance of point clouds, validating our method's effectiveness when using high-quality normal inputs.
We further train and evaluate our method using depth-computed normals ~\cite{qi2018geonet} or cost-computed normals \cite{kusupati2020normal}, the results are suboptimal as they essentially rely on the quality of input depth, which can degrade in challenging areas.
Though lacking multi-view consistency, monocular normals do not collapse in challenging geometric estimation regions of the cost volume. 
This reveals a nice property for monocular estimations. In addition to the DTU dataset, we also observe notable improvement using monocular surface normals on other benchmarks.

\begin{table}[t]
\centering
 \renewcommand{\arraystretch}{1.1}
 \setlength{\tabcolsep}{7pt}
\scalebox{0.8}{
\begin{tabular}{lccc}
\toprule
Aggregation kernel $(d\times h\times w)$         & Acc. $\downarrow$           & Comp. $\downarrow$        & Overall $\downarrow$        \\ 
\midrule
Standard Conv3D$ (3\times 3\times 3)$     & 0.365          & 0.265          & 0.315          \\
Standard Conv3D $(5\times 3\times 3)$   & 0.352          & 0.260          & 0.306          \\
Standard Conv3D $(7\times 3\times 3)$    & 0.352          & 0.258          & 0.305          \\
Proposed GCA $(3\times 3\times 3)$   & \textbf{0.347} & \textbf{0.227} & \textbf{0.287} \\ 
\bottomrule
\end{tabular}}
\caption{\textbf{Evaluation of aggregation receptive fields.}
Directly expanding receptive fields along the depth dimension yields limited improvement and is easily saturated. In contrast, our method achieves the best performance with a kernel size of 3.}%
\label{table::dif_ganshouye}
\end{table}

\begin{table}[t]
\centering
 \renewcommand{\arraystretch}{1}
 \setlength{\tabcolsep}{14pt}
\scalebox{0.73}{
\begin{tabular}{lccc}
\toprule
Method          & Acc. $\downarrow$           & Comp. $\downarrow$          & Overall $\downarrow$        \\ 
\midrule
Ours + Depth-to-normal~\cite{qi2018geonet} & 0.352          & 0.242          & 0.297          \\
Ours + Cost-to-normal~\cite{kusupati2020normal}  & 0.358          & 0.241          & 0.300          \\
Ours + Mono-normal~\cite{eftekhar2021omnidata}     & 0.347 & 0.227 & 0.287 \\ 
\midrule
Ours + GT normal    & \textbf{0.275} & \textbf{0.221} & \textbf{0.248} \\ 
\bottomrule
\end{tabular}}
\caption{\textbf{Evaluation of different normal cues.}
Our method with GT normal demonstrates remarkable performance (0.248). Among all estimated normals, the off-the-shelf monocular normal has the best performance.
}
\vspace{-5pt}
\label{table::dif_normal}
\end{table}

\section{Conclusion}
In this paper, we propose GoMVS, which aggregates locally consistent geometries to better utilize adjacent geometry. By leveraging local smoothness in conjunction with surface normal, we propose geometrically consistent aggregation. it computes the correspondence from the adjacent depth hypotheses space to the reference depth space and propagates cost accordingly. 
Furthermore, we investigate different choices for generating normal priors and find that monocular cues effectively complement the MVS network.
Our method achieves state-of-the-art performance on the DTU, Tanks and Temples, and ETH3D datasets.
\par
{\noindent \textbf{Acknowledgements}
Y. Zhang was supported by NSFC (No.U19B2037) and the Natural Science Basic Research Program of Shaanxi (No.2021JCW-03). Y. Zhu was supported by NSFC (No.61901384).}
{
    \small
    \bibliographystyle{ieeenat_fullname}
    \bibliography{main}
}


\end{document}